# Comparative analysis of subword tokenization approaches for Indian languages

Sudhansu Bala Das[a],∗ , Samujjal Choudhurya , Tapas Kumar Mishra[a] , Bidyut Kr. Patra[b]

[a]*National Institute of Technology(NIT), Rourkela, Odisha, India*

[b] *Indian Institute of Technology (IIT), Varanasi, Uttar Pradesh, India*

**Abstract**

Tokenization is the act of breaking down text into smaller parts, or tokens, that are easier for machines to process. This is a key phase in machine translation (MT) models. Subword tokenization enhances this process by breaking down words into smaller subword units, which is especially beneficial in languages with complicated morphology or a vast vocabulary. It is useful in capturing the intricate structure of words in Indian languages (ILs), such as prefixes, suffixes, and other morphological variations. These languages frequently use agglutinative structures, in which words are formed by the combination of multiple morphemes such as suffixes, prefixes, and stems. As a result, a suitable tokenization strategy must be chosen to address these scenarios. This paper examines how different subword tokenization techniques, such as SentencePiece, Byte Pair Encoding (BPE), and WordPiece Tokenization, affect ILs. The effectiveness of these subword tokenization techniques is investigated in statistical, neural, and multilingual neural machine translation models. All models are examined using standard evaluation metrics, such as the Bilingual Evaluation Understudy (BLEU) score, TER, METEOR, CHRF, RIBES, and COMET. Based on the results, it appears that for the majority of language pairs for the Statistical and Neural MT models, the SentencePiece tokenizer continuously performed better than other tokenizers in terms of BLEU score. However, BPE tokenization outperformed other tokenization techniques in the context of Multilingual Neural Machine Translation model. The results show that, despite using the same tokenizer and dataset for each model, translations from ILs to English surpassed translations from English to ILs.

**Keywords:** *Machine Translation, Indic Languages, Tokenization, Evaluation Metrics*

## 1. Introduction

Language is a constantly evolving medium for communicating ideas and emotions. It not only serves as a standard of expressing our emotions and thoughts, but also takes communication a step further by providing context and cultural significance in addition to words. Since languages differ from region to region, intercommunication among people becomes challenging in the absence of interpreters. This is when natural language processing (NLP) arrives into play. NLP is a branch of artificial intelligence that investigates the relationship between machines and human language. The overall objective is to allow machines to comprehend, translate, and produce human-like text, allowing for more effective interactions between people who speak different languages. Machine Translation (MT) arises as an important application in the field of NLP. At its core, machine translation (MT) uses algorithms for computation to automatically convert text or speech between languages. The importance of MT stems not only from the way it can help interaction between cultures, but also from its capacity to bring down linguistic obstacles in real-time, encouraging international communication, awareness, and sharing of information [13]. As MT evolves, the difficulties it encounters become more diverse and complex. Language complexities, such as sentence structures, idiomatic phrases, and cultural differences, present significant challenges in producing precise and relevant context translations. In the field of MT, statistical machine translation (SMT) has served an important role in breaking down obstacles associated with language [14]. SMT uses statistical models developed using large datasets to generate probabilistic predictions regarding text translation from one language to another. Unlike rule-based systems [15], which use predetermined rules of grammar, SMT acquires patterns through large corpora, enabling it to adjust to the complexity of natural language. The mathematical framework of SMT enables the system to make choices according to the likelihood of particular translation choices [16]. The statistical method improves SMT's adaptability, rendering it efficient over a wide range of language pairs and subject areas. In the realm of translation problems, SMT faces significant challenges in successfully handling and comprehending input text. Despite this, the adaptability of these methods to new languages or domains is limited, as well as the creation and upkeep of rules or models may necessitate a significant amount of manual labor. On the other hand, Neural Machine Translation (NMT) is a paradigm shift that uses neural networks to directly learn the translation mapping from input to output sequences [17] [47]. In contrast to SMT, NMT models translate more fluently being contextually accurate because of their encoder-decoder architecture, which captures complex relationships among words and phrases. Furthermore, advances in NMT have resulted in the development of Multilingual Neural Machine Translation (MNMT) [18]. Researchers' interest in multilingual neural machine translation (MNMT) has grown over the past few years. One of the most significant benefits of MNMT is the ability to translate multiple language pairs using only one model [19]. While SMT, NMT, and MNMT have different underlying architectures and techniques, they all use tokenization as a preprocessing step. Tokenization is the process of breaking down input text into smaller units, that include words or subwords so that translation models can process them more efficiently. Proper tokenization is critical for accurately capturing linguistic nuances in Indian languages (ILs), which are known for their complicated structure and diverse morphological features. Specialized tokenization techniques designed for the unique characteristics of ILs can help enhancing translation precision as well as fluency [22]. Exploring and implementing appropriate tokenization methods is therefore necessary for producing high-quality machine translations for Indian languages. Tokenization aids in the identification of specific morphemes inside words, which are necessary for comprehending the structure as well as the significance of words in the framework of morphological analysis [24]. By properly tokenizing the text, we can distinguish suffixes, prefixes, root phrases, and various other morphological components, enabling more accurate interpretation and analysis [23]. Hence, tokenization is an essential component of the morphological analysis process, as it allows for the detection and evaluation of distinct morphemes within words, which is required for machine translation tasks. Hence, the main objective of this paper is to analyze and compare the efficacy of various subword tokenization methods in the scenario of statistical, neural, and multilingual models for 11 Indian languages. The investigation focuses on three different subword tokenization methods: Byte Pair Encoding (BPE) [9], WordPiece [25], and SentencePiece [21]. These methods have been chosen due to their applicability and possible influence on machine translation efficiency. The evaluation of these techniques attempts to identify the advantages and disadvantages of each for handling text data written in Indian languages. This paper aims to offer important insights into the best tokenization strategy for Indian language processing tasks, especially in the context of machine translation (MT), through experimentation and analysis.

**Hence, the main contributions of this paper are as follows:**

1. As a first attempt ever, to the best of our knowledge, this paper examines the effects of different subword tokenization methods, such as WordPiece [25], SentencePiece [21], and Byte Pair Encoding (BPE) [9], on 11 Indian languages.

2. The paper explores the significance of tokenization in various machine translation models, including SMT, NMT, and MNMT, and provides comparisons as well as insights into their efficacy.

3. The paper compares the efficacy of different MT models as well as tokenization methods across 11 Indian languages using a wide range of MT metrics such as BLEU [1], RIBES [2], METEOR [3], TER [4], CHRF[5] and COMET [6].

The remaining part of this paper is organized as follows: Section 2 describes and gives a brief overview of the tokenization and some notable work related to it. Section 3 discusses the different tokenization and their impact on Statistical, Neural, and Multilingual Models. Section 4 describes the experimental framework, and Section 5 presents a few evaluation metrics. Section 6 discusses the results and Section 7 presents conclusions.

## 2. Tokenization

Early studies emphasize the significance of utilizing a tokenization algorithm since tokenization constitutes one of the first phases in any information retrieval or natural language processing system. Over the last few decades, tokenization has undergone numerous phases. One of the old tokenization techniques is whitespace tokenization, which divides text into sections using whitespace characters like tabs and spaces. Even though whitespace tokenization is easy to use and straightforward, it might not be adequate for tasks requiring more precise token granularity or for languages with intricate morphological structures.Traditional tokenization methods divide text into words according to whitespace or punctuation marks [29]. However, this method may result in out-of-vocabulary (OOV) words and impair translation quality in languages with complex morphology or huge vocabularies. More advanced methods, like word-based tokenization [28], have been developed by researchers to overcome the drawbacks of whitespace tokenization. Word tokenization algorithms, which frequently rely on dictionaries or language specific rules, seek to divide text into meaningful word units. The NLP community, for instance, makes extensive use of the word tokenize function in the NLTK library [26] and the word tokenize method in the spaCy library [27]. Several notable studies in the literature concentrate on improving these subword tokenization techniques. Ding et al [30], for instance, investigate how the quantity of BPE merges affects the effectiveness of machine translation. Provilkov et al [31] have suggest a drop-out technique for each merge step of BPE that enhances machine translation effectiveness by breaking the deterministic nature of the algorithm. According to Bostrom and Durrett [32], there are better methods for language pretraining than BPE because it does not make good use of the vocabulary space. To evaluate the impact of incorporating subword tokens, Nayak et al. [33] compare the activations of BERT's attention layers with WordPiece and word-level tokenization. They discover that modeling semantically significant connections among words is hampered by a vocabulary that relies heavily on frequency-based character combinations. Furthermore, representations derived from tokenization utilizing word occurrence data rely more on frequency data than semantics [34]. Banerjee et al [35] have integrated BPE with a commercially available morphological segmenter in translations from Bengali and Hindi against English. In addition to statistical segmentation techniques, employed a retrained version of the linguistically inspired segmentation model for Arabic. Recent developments in deep learning have further extended the range of tokenization techniques. Subword tokenization as well as learned embeddings are used by models such as BERT [37] and GPT [38] to effectively gather contextual information. By integrating contextualized embeddings, such models may more accurately capture the nuances and complexities of natural language. It reduces the complexity of language data, which facilitates processing and analysis by algorithms. It also balances text representation, guaranteeing coherence among various language processing models and tasks. According to Park et al [30], BPE is not the best language method for pretraining because it does not make good use of the vocabulary space. To evaluate the impact of incorporating subword tokens, Nayak et al [41] contrast the activations of BERT's attention layers with WordPiece and word-level tokenization. They discover that modeling semantically significant connections between words is hampered by a vocabulary that relies heavily on frequency-based character combinations.

## 3. Different Tokenization and their Impact on Statistical, Neural and Multilingual Models

In this section, different subword tokenizations used for our experiments are discussed. Subword tokenization is an approach that is especially important for Indian languages (ILs) because of their complex morphology and lack of a standardized orthography. The term "complex morphology" refers to how words in a language can change forms depending on the situation. For example, in languages like Bengali or Hindi, words can have different forms depending on factors like number, gender, and tense. Subword tokenization algorithms such as Byte-Pair Encoding (BPE) [9], SentencePiece [21], and WordPiece [25] have successfully handled the complexity of languages like Hindi, Bengali, and Tamil by breaking down words into smaller parts. For breaking down the text into subword units utilizes a range of linguistic characteristics, including grammatical rules, character sequences, and frequency. This division not only helps to capture the language's morphological complexity but also makes it easier to handle out-of-vocabulary (OOV) words [42]. When a model encounters an unexpected word, subword tokenization allows it to interpret the word as an ordered set of subword units from its lexicon, increasing its capacity to process and evaluate different text data. This makes it easier to create translation systems that are more precise and suitable for the context, which helps to overcome communication barriers and provides wider access to resources and information across linguistic divides. As research in this area continues, it is anticipated that the accuracy and availability of machine translation for Indian languages will improve, contributing to their preservation, outreach, and broad participation in the digital age. However, it is still unclear which subword tokenization method needs to be used to determine what is best for Indian languages (ILs).

## 4. Experimental Framework

### 4.1. Dataset

The training dataset for Indic languages(ILs) is taken from Samanantar, a large publicly accessible dataset for ILs [7].This dataset contains more than 49.6 million sentence pairs translated between English and 11 ILs. The Flores200 dataset is used for testing purposes [8]. These datasets are used to evaluate the efficacy of three tokenizers, i.e., BPE [9], SentencePiece [21], and WordPiece, and their impact on ILs using SMT (Statistical Machine Translation), NMT (Neural Machine Translation), and MNMT (Multilingual Neural Machine Translation) models are investigated.

Table 1: Dataset Statistics

| English to Indic | Parallel Dataset |
|---|---|
| Tamil(TA) | 5.16M |
| Assamese (AS) | 0.14M |
| Marathi(MR) | 3.32M |
| Malayalam (ML) | 5.85M |
| Telugu(TE) | 4.82M |
| Bengali(BN) | 8.52M |
| Gujarati(GU) | 3.05M |
| Hindi(HI) | 8.56M |
| Kannada(KN) | 4.07M6 |
| Odia(OR) | 1.00M |
| Punjabi(PA) | 2.42M |

### 4.2. Preprocessing

A few punctuation marks in the extended Unicode have been changed to their standard equivalents. The characters with accents were eliminated. Numbers in Indian languages were converted from English to Indic scripts. The corpus was cleaned up by eliminating unprintable characters, extra spaces, and characters that did not belong in the standard alphabets of the language pair. Finally, the corpus was cleaned up to remove any redundant quotation marks. Further Preprocessing is performed to effectively handle diacritics such as halants and nuktas and change all the text in the English language datasets to lowercase. For example, in Tamil எந்

களிடம ்ॖ இப ்ॖ ்போது is changed into எங்களிடம்இப்்போது .
Additional preprocessing, such as changing all text in the language datasets to lowercase.

**4.3. Different Tokenization and their Impact on Statistical, Neural and Multilingual Models Byte Pair Encoding (BPE):**

**Byte-pair encoding (BPE)** is a compression tokenization utilized in machine translation (MT) for representing a large vocabulary using a small number of subword units [9]. The utilization of byte-pair encoding makes sure that those that are the most common words within the vocabulary will appear as single tokens, while the rarest words are divided into subword tokens. An example of a sentence tokenized using BPE is:

**Original**

यह खोज पक्षियों में पंखों के विकास की पूरी जानकारी भी प्रदान करती है.

**Segmented Sentence**

यह खोज पक ्ॖ षियों में पं@@ खों के विकास की पूरी जानकारी भी प ्ॖ रदान करती है .

**Algorithm 1** Byte Pair Encoding (BPE)

1: **Initialize Vocabulary:** Start with a text dataset.
2: **Split Words:** Each word in the dataset should be represented as a string of characters with an additional special end-of-word token.
3: **Merge Frequent Pairs:** Determine which character pairs appear together the most frequently, then combine them into a single token.
4: **Update Vocabulary:** Iteratively repeat the merging process, using the merged tokens, and update the vocabulary.
5: **Stop Criteria:** Stop while all iterations are completed or a predetermined token limit size has been reached.

It plays an important role in statistical machine translation (SMT), neural machine translation (NMT), and multilingual neural machine translation (MNMT) models. MOSES is an open-source toolkit for SMT. The first step for constructing the SMT model is to preprocess the data. In this step, the text data undergo multiple transformations in the initial preprocessing stage before the SMT model is constructed. To start, all language data is changed to lowercase to maintain corpus consistency. Similarly, to prevent encoding problems and simplify the dataset, accented characters are eliminated from the text. Even for uniformity's sake, some punctuation marks in the extended Unicode are also changed to their standard counterparts. After that, the corpus is carefully cleaned to remove extraneous spaces, unwanted characters, and nonstandard alphabets unique to the language pair. Then, a few more processes are done for Indian languages, allowing them to efficiently deal with diacritics such as halants and nuktas [43]. With this change, the SMT model is ensured to maintain the linguistic subtleties of Indian languages (ILs). After that, Byte Pair Encoding (BPE) tokenization is applied to the data. For BPE, the subword-nmt is used. In SMT, BPE initially tokenizes the source and target language data into subword units, effectively handling unusual or out-of-vocabulary phrases. By splitting words into smaller parts, BPE ensures the model can acquire significant representations for less common keywords, enhancing translation accuracy. Following the completion of the tokenization process in Statistical Machine Translation (SMT), the next steps include language model training and translation system training. During language model training, a model is created to ensure fluency in output translation. Utilizing the target language data, this model is constructed. The language and translation models are trained on the training dataset and binarized. In Moses, the training method uses word and segment occurrences to connect the target and source languages. Using the training dataset, the language and translation models are binarized after training. After training the language model, the next step is to train translation models. The final model is filtered using the test dataset before being utilized for translating the preprocessed test dataset from the source to the target language. Once training was done, the translation file was detokenized using the Moses detokenizer, and superfluous quote marks were eliminated.

Similarly, with NMT, BPE tokenization divides words into subword units, allowing the model to handle text more effectively. This method enables the model to learn from a larger set of subword units and to understand detailed linguistic patterns and nuances in both the source and target languages. The Fairseq library [40] is used for building NMT systems. All NMT models are implemented using the transformer model [39]. The model consists of Six encoder and six decoder layers. The encoder and decoder each have 8 attention heads. The dimensions of the Transformer feed-forward layer are 2048. The embedding dimensions of the decoder and encoder are 512.

In MNMT, BPE is critical for handling multilingual datasets. By breaking down text into subword units, BPE allows the model to interpret many languages at the same time. Because of this flexibility, MNMT models can handle a wide range of language pairs and produce appropriate translations in various linguistic circumstances. The MNMT model is constructed using the same configuration that is utilized to build the NMT model utilizing the Transformer model from Fairseq[40].

**WordPiece**:

Like BPE, WordPiece begins with a vocabulary made up of individual characters or bytes, which it then progressively combines into larger subword units according to how frequently those characters appear in the training set [25]. It joins consecutive pairs of whole words or subword units rather than consecutive pairs of bytes. An example of a sentence tokenized using WordPiece is:

**Original**

यह खोज पक्षियों में पंखों के विकास की पूरी जानकारी भी प्रदान करती है.

**Segmented Sentence**

_यह _खोज _पक _ㅇ _षियों _में _प ंख ों _के _विकास _की _पूरी _जानकारी _भी _प _ㅇ _रदान _करती _है _.

Using the base vocabulary as a training set, the WordPiece tokenization algorithm selects the pair with the highest likelihood, adds it to the vocabulary, trains the language model using the new vocabulary, and repeats the process until the needed vocabulary size or likelihood threshold is met. In WordPiece, the merge decision depends on a count metric estimated as follows:

$$Count(x, y) = \frac{frequency of(x, y)}{frequency(x) \times frequency(y)}$$

This count shows how often the two symbols (x, y) occur in the sum of the frequencies of the individual symbols, x and y. The symbols that have the highest count together are chosen to be combined into the vocabulary. For wordpiece, huggingface tokenizers library is utilized. It serves as an important preprocessing step in Statistical Machine Translation (SMT) for translating English to ILs and vice versa. It tokenizes sentences into subword units, ensuring that even uncommon or non-vocabulary terms are correctly represented. Similarly, in Neural Machine Translation (NMT) experiments, wordpiece tokenization comes after preliminary preprocessing, which typically includes steps like lowercase and normalisation. It is used in NMT to separate text into subword units rather than full words. By representing uncommon or non-vocabulary terms with significant subword units, this tokenization helps handle them and improves the model's ability to analyze and make generalizations from the input data. Once tokenization has been completed, a dictionary is created to map every single subword unit to a distinct identifier, allowing more efficient data processing. The preprocessed data is then subjected to binarization, transforming the textual data into a numerical format appropriate for training the model. The transformer model [37] is used to train the NMT models. The model comprises six encoder-decoder layers that are optimized using the Adam optimizer[17]. Likewise, the MNMT(Multilingual Neural Machine Translation) model uses WordPiece tokenization, which divides each word in the input text into subword units determined by the vocabulary learned while on tokenization. The segmentation process enables the model to deal with various linguistic variations throughout multiple languages by using essential subword units to represent uncommon or out-of-vocabulary terms. Then, the model is trained using the Transformer model[37] from Fairseq[25], where the configuration utilized is the same as that of the NMT model.

**SentencePiece**:

Traditional tokenization algorithms have difficulty when dealing with languages that do not utilize spaces for separating words. As these algorithms usually consider a word-space structure in the input text, this presents a significant challenge. However, SentencePiece solves this issue by addressing the input as a raw input stream with spaces as part of the character set [19]. This novel approach allows SentencePiece to deal with languages with diverse word boundaries more efficiently. It does word segmentation regardless of language tokenization conventions by treating spaces as regular characters within the input text. An example of a sentence tokenized using SentencePiece is:

Original Sentence: यह खोज पक्षियों में पंखों के विकास की पूरी जानकारी भी प्रदान करती है.

Segmented SEntence: ▁यह ▁खोज ▁पक ् ▁षियों ▁में ▁प ं ▁ख ों ▁के ▁विकास ▁की ▁पूरी ▁जानकारी ▁भी ▁प ् र ▁दान ▁करती ▁है ▁.

It is a flexible subword tokenization algorithm created by Google that uses unsupervised neural network approaches. It analyzes the input sentences directly at the subword level, dissecting them into a series of variable length subword units, in contrast to BPE and WordPiece. It divides sentences into significant subword units by applying a neural network model and a predefined vocabulary of subword units.

---
**Algorithm 2** Training SentencePiece
---
1: **Input:** Dataset, Vocabulary size, Model type (subword/character)
2: Initialize an empty vocabulary.
3: Tokenize the input text into subwords or characters.
4: Count the frequency of subword sequences in the tokenized text.
5: Sort the subword sequences based on frequency.
6: Select the top $N$ most frequent subword sequences to form the vocabulary, where $N$ is the desired vocabulary size.
7: **Output:** Constructed vocabulary.

SentencePiece plays an important role in tokenizing text data for preprocessing in Statistical Machine Translation (SMT), Neural Machine Translation (NMT), and Multilingual Neural Machine Translation (MNMT). For SentencePiece tokenization, the SentencePiece library is utilized. It helps for preprocessing data in both source and target languages in SMT, ensuring that significant linguistic variations and patterns are captured during the tokenization process. Similarly, it is also used in NMT to tokenize input sentence sequences, allowing the model to accommodate a variety of languages and linguistic conventions.

Even the SentencePiece helps in MNMT by preprocessing bilingual corpora sets, concatenating them to facilitate training, and creating proper vocabularies for multilingual translation tasks. All of our experiments on the use of tokenizers in Statistical Machine Translation (SMT), Neural Machine Translation (NMT), and Multilingual Neural Machine Translation (MNMT) models show consistent configuration. This consistency ensures dependability and comparable results when comparing the effectiveness of various translation frameworks.

## 5. Results and Discussion

### 5.1. Impact on Translation using Different Tokenizations

While analyzing the translation using different tokenizers, it has been observed that the translation quality is impacted based on the token splits of the text. For example, while translating the English sentence, "He said, now we have 4 months old rats who are not diabetic who were diabetic" to Hindi language using different tokenizers such as WordPiece, SentencePiece, and Byte Pair Encoding (BPE).

**WordPiece**: It is a token segmented based on its probability and frequency in the vocabulary:
Tokens: [उन्होंने, कहा, „, अब, हमारे, पास, 4, महीने, के, चूहे, हैं, जो, मधु, ## मेह, ग्रस्त, नहीं, हैं, „ लेकिन, पहले, मधु, ## मेह, ग्रस्त, थे, ▢]
Here, "मधुमेह" is segmented into "मधु" and "## मेह" depending on their frequency and probability. Other words, like "उन्होंने" and "कहा", remain whole as they frequently appear in the training data.

**SentencePiece**: It utilizes the entire input sentence, considering it as one string without initial segmentation:
Tokens: [▢उन्होंने, ▢कहा, ▢„, ▢अब, ▢हमारे, ▢पास, ▢4, ▢महीने, ▢के, ▢चूहे, ▢हैं, ▢जो, ▢मधुमेह, ▢ग्रस्त, ▢नहीं,▢हैं, ▢„ ▢लेकिन, ▢पहले, ▢मधुमेह, ▢ग्रस्त, ▢थे, ▢▢]
Here, sentencePiece keeps words like "मधुमेह" and "ग्रस्त" as whole tokens since they likely appear frequently in the corpus.

**Byte Pair Encoding (BPE)**: It segments the tokens based on frequently occurring character pairs, focusing on morphology:
Tokens: [उन्होंने, कहा, „, अब, हमारे, पास, 4, महीने, के, चूहे, हैं, जो, मधु, +मेह, ग्र, +स्त, नहीं, हैं, „ लेकिन, पहले, मधु, +मेह, ग्र, +स्त, थे, ▢]
Likewise, BPE segments "मधुमेह" into "मधु+मेह" and "ग्रस्त" into "ग्र+स्त," which indicated the common morphological structure of Hindi words.

## 5.2 Analysis of Result

Tables 2, 3, and 4 display all evaluation metrics results using various tokenizations with different MT evaluation metrics. MT utilizes a range of measures to assess system performance since each captures a unique component of translation quality. All models utilized for our experiments use Flores200 test sets. Fairseq library with Adam optimizer with betas of (0.9,0.98) for training is used. Our model is run on a high-performance workstation equipped with an Intel Xeon W-1290 CPU, with 10 physical cores and 20 threads (3.20 GHz base frequency, up to 5.20 GHz boost), providing robust multi-threading and caching with 20 MiB of L3 cache. The system includes 62 GB of RAM and an NVIDIA Quadro RTX 5000 GPU with 16 GB of VRAM, supported by driver version 535.154.05. The system uses CUDA 11.5 for compilation and is compatible with CUDA 12.2 for runtime operations, optimizing model training performance. Training each MNMT model takes roughly two and a half days,

whereas training for SMT and NMT models took from half to two days, according to its data size.

MT evaluation metrics are a process that requires not only linguistic accuracy but also fluency, coherence, and semantic alignment with the source text. Each evaluation metric has its own advantage and working principle, which motivates us to check the translation quality. Each metric emphasizes a particular aspect of translation quality, such as semantic accuracy (COMET, METEOR) or structural accuracy or fluency (RIBES, CHRF). Bilingual Evaluation Understudy (BLEU) is one of the famous measures for MT evaluation, and it evaluates the overlap of n-grams (word sequences) among the machine translation and one or more reference translations [1]. It highlights and captures exact matches of words and phrases, giving it an accurate indicator of surface-level similarity across generated and reference translations. The popularity of BLEU derives from its simplicity and computing efficiency. For all the experiments, the BLEU score ranges from 0 to 100. Higher BLEU scores usually suggest translations that closely resemble human references. Similarly, the Translation Edit Rate (TER) determines the number of edits needed to convert a machine translation into an exact match to the reference translation [4]. Insertions, deletions, substitutions, and shifts are some examples of edits. It can capture and measure the effort required to make a machine translation human-like, demonstrating how much post-editing is required to appear natural. TER scores vary between 0 and 1, with lower scores implying higher translation quality (i.e., fewer edits needed to match the reference translation) and higher scores indicating lower quality. Meanwhile, Metric for Evaluation of Translation with Explicit ORdering (METEOR) tries to improve upon BLEU by incorporating synonyms, partial matches, stemming, and word order [3]. It assigns varying weights to various types of matches. It captures semantic similarity beyond direct word matches by accounting for synonyms and associated words, resulting in a more adaptable measure of meaning overlap. METEOR, with its support for synonyms and stemming, fits more closely with human judgment on translation adequacy and fluency. However, CHRF analyzes character n-gram matches instead of word n-grams, concentrating on subword similarity [5]. It computes the F-score by balancing precision and recall for these character-level n-grams. It is susceptible to slight spelling or morphological variations, and it works best with morphologically rich and highly inflected languages. CHRF's emphasis on character similarity may punish translations that rephrase at the sentence level or utilize synonymous phrases, ignoring broader semantic accuracy. The Rank-based Intuitive Bilingual Evaluation Score (RIBES) is intended to assess translation quality with a focus on word order [2]. It assesses the connection of word places in the source and translated sentences. It captures and highlights word order along with syntactic coherence; RIBES preserves structural accuracy, which is critical in translations where word order strongly affects meaning. Similarly, the Cross-lingual Optimized Metric for Evaluation of Translation (COMET) is a neural-based evaluation metric that uses embeddings from pre-trained language models to determine translation quality [6]. It utilizes a regression model trained to estimate human scores by analyzing source and target language translations. It captures the deep semantic alignment among source and target texts, recognizing semantic similarities and faults across languages. To compute the BLEU score, TER, CHRF, and SacreBleu [45] are utilized, whereas the HuggingFace library [46] is used to evaluate the rest of the metrics.

Table 2 shows that using Byte Pair Encoding, MNMT models outperformed SMT and NMT models. Using BPE tokenization, MNMT's BLEU score falls between 5.14 and 32.19 whereas the TER score falls between 56.84 and 102.9. The CHRF ranges from 35.81 to 62.35 and the METEOR score is between 0.24 and 0.63. COMET scores lie between 0.73 to 0.87 and the RIBES score is between 0.34 and 0.79 as shown in Table 2. In SMT, BLEU scores range from 1.3 to 10.19, and TER scores from 81.1 to 118.10. The range of the CHRF score is 26.07 to 46.23. COMET scores fall between 0.49 and 0.79, while METEOR scores span from 0.06 to 0.35. Furthermore, the range of RIBES scores is 0.13 to 0.58. However, for NMT, the ranges of scores are as follows: BLEU scores lie between 0.26 to 32.59 and TER scores range from 57.82 to 178.79, whereas METEOR ranges from 0.07 to 0.61. CHRF from 9.55 to 61.89, COMET from 0.51 to 0.86, and RIBES from 0.144 to 0.78. While comparing the results, it is observed that for MNMT models, the EN-AS model has the lowest BLUE score and EN-HI model achieves the highest BLEU score. The reason could be the smaller corpus of AS, which results in lower translation quality than other languages. At the same time, the combination of qualitative and quantitative data in HI leads to better performance as compared to other languages. MNMT model can handle translation tasks for language pairs despite limited parallel data. It performs better as it can exchange knowledge and enhance the quality of translation for all involved languages by working together to train in multiple languages. It learns common representations throughout languages, allowing it to use similarities between languages to enhance translation quality, particularly for low-resource languages.

Table 2: Evaluation Metrics using BPE as Tokenization

| Model | Language Pair | BLEU | TER | METEOR | CHRF | RIBES | COMET |
|---|---|---|---|---|---|---|---|
| SMT | EN-AS | 1.6 | 95.16 | 0.14 | 26.07 | 0.49 | 0.6 |
| | AS-EN | 3.35 | 87.63 | 0.24 | 30.06 | 0.46 | 0.68 |

|  | EN-ML | 1.3 | 99.12 | 0.08 | 28.7 | 0.09 | 0.67 |
|---|---|---|---|---|---|---|---|
|  | ML-EN | 4.29 | 85.87 | 0.25 | 32.06 | 0.42 | 0.71 |
|  | EN-BN | 3.06 | 88.25 | 0.16 | 28.58 | 0.51 | 0.76 |
|  | BN-EN | 10.19 | 81.74 | 0.37 | 42.25 | 0.55 | 0.73 |
|  | EN-MR | 2.1 | 92.82 | 0.13 | 26.31 | 0.37 | 0.7 |
|  | MR-EN | 6.7 | 84.03 | 0.31 | 37.15 | 0.511 | 0.56 |
|  | EN-GU | 3.68 | 88.87 | 0.19 | 27.86 | 0.48 | 0.76 |
|  | GU-EN | 8.4 | 83.83 | 0.35 | 40.66 | 0.53 | 0.76 |
|  | EN-KN | 2.13 | 94.43 | 0.13 | 30.23 | 0.27 | 0.7 |
|  | KN-EN | 6.53 | 84.63 | 0.3 | 37 | 0.48 | 0.7 |
|  | EN-HI | 6.9 | 82.38 | 0.25 | 29.27 | 0.54 | 0.49 |
|  | HI-EN | 1.4 | 96.88 | 0.06 | 8.35 | 0.19 | 0.63 |
|  | EN-OR | 3.67 | 94.16 | 0.22 | 36.24 | 0.55 | 0.72 |
|  | OR-EN | 7.15 | 84.66 | 0.36 | 39.56 | 0.52 | 0.74 |
|  | EN-PA | 5.21 | 83.9 | 0.241 | 28.83 | 0.58 | 0.74 |
|  | PA-EN | 8.68 | 81.1 | 0.35 | 40.09 | 0.54 | 0.71 |
|  | EN-TE | 2.7 | 94.4 | 0.14 | 30.02 | 0.3 | 0.72 |
|  | TE-EN | 8.09 | 84.06 | 0.34 | 39.66 | 0.5 | 0.71 |
|  | EN-TA | 2.99 | 118.1 | 0.16 | 46.23 | 0.13 | 0.73 |
|  | TA-EN | 9.09 | 87.22 | 0.38 | 42.67 | 0.5 | 0.79 |
| **NMT** | EN-AS | 0.26 | 135.15 | 0.07 | 9.55 | 0.14 | 0.54 |
|  | AS-EN | 0.77 | 178.79 | 0.17 | 20.5 | 0.18 | 0.51 |
|  | EN-ML | 8.11 | 106.55 | 0.29 | 52.91 | 0.44 | 0.83 |
|  | ML-EN | 22.13 | 71.31 | 0.55 | 53.62 | 0.71 | 0.83 |
|  | EN-BN | 3.06 | 88.25 | 0.16 | 28.58 | 0.51 | 0.76 |
|  | BN-EN | 28.22 | 62.6 | 0.61 | 58.03 | 0.76 | 0.84 |
|  | EN-MR | 9.51 | 97.43 | 0.34 | 44.71 | 0.6 | 0.81 |
|  | MR-EN | 19.37 | 73.42 | 0.51 | 50.21 | 0.7 | 0.69 |
|  | EN-GU | 16.29 | 81.43 | 0.42 | 49.41 | 0.67 | 0.84 |
|  | GU-EN | 23.75 | 70.05 | 0.57 | 55.3 | 0.73 | 0.85 |
|  | EN-KN | 11.86 | 89.79 | 0.34 | 52.15 | 0.58 | 0.82 |
|  | KN-EN | 20.84 | 74.33 | 0.53 | 52.92 | 0.71 | 0.82 |
|  | EN-HI | 31.41 | 57.82 | 0.56 | 56.6 | 0.78 | 0.86 |
|  | HI-EN | 32.59 | 57.66 | 0.65 | 61.89 | 0.79 | 0.79 |
|  | EN-OR | 5.09 | 99.14 | 0.24 | 36.58 | 0.58 | 0.75 |
|  | OR-EN | 10.92 | 84.95 | 0.38 | 39.27 | 0.58 | 0.75 |
|  | EN-PA | 19.16 | 73.58 | 0.48 | 48.53 | 0.74 | 0.84 |
|  | PA-EN | 27.39 | 61.92 | 0.59 | 56.31 | 0.77 | 0.81 |
| | EN-TE | 13.73 | 91.8 | 0.39 | 54.41 | 0.61 | 0.83 |
| | TE-EN | 24.52 | 68.95 | 0.56 | 55.36 | 0.73 | 0.82 |
| | TA-EN | 7.03 | 107.93 | 0.24 | 52.64 | 0.31 | 0.81 |
| | EN-TA | 20.99 | 74.36 | 0.53 | 52.32 | 0.71 | 0.84 |
| **MNMT** | EN-AS | 5.14 | 93.13 | 0.24 | 35.81 | 0.61 | 0.8 |
| | AS-EN | 20.51 | 72.4 | 0.51 | 50.52 | 0.71 | 0.78 |
| | EN-ML | 9.71 | 94.77 | 0.3 | 54.97 | 0.48 | 0.85 |
| | ML-EN | 27.76 | 64.39 | 0.6 | 58.55 | 0.75 | 0.85 |
| | EN-BN | 16.11 | 73.34 | 0.42 | 53.03 | 0.72 | 0.86 |
| | BN-EN | 29 | 61.83 | 0.62 | 58.7 | 0.77 | 0.85 |
| | EN-MR | 11.01 | 88.57 | 0.36 | 47.66 | 0.64 | 0.84 |
| | MR-EN | 27 | 62.63 | 0.59 | 57.2 | 0.76 | 0.72 |
| | EN-GU | 18.23 | 75.07 | 0.45 | 51.82 | 0.72 | 0.87 |
| | GU-EN | 30.93 | 59.35 | 0.63 | 60.67 | 0.78 | 0.86 |
| | EN-KN | 12.11 | 86.11 | 0.35 | 53.17 | 0.6 | 0.84 |
| | KN-EN | 25.5 | 67.65 | 0.58 | 56.08 | 0.74 | 0.83 |
| | EN-HI | 29.89 | 59.43 | 0.55 | 56.17 | 0.78 | 0.87 |
| | HI-EN | 33.31 | 56.85 | 0.65 | 62.35 | 0.79 | 0.79 |
| | EN-OR | 11.21 | 79.5 | 0.353 | 48.38 | 0.69 | 0.85 |
| | OR-EN | 26.91 | 63.78 | 0.6 | 57.36 | 0.75 | 0.81 |
| | EN-PA | 20.82 | 68.43 | 0.5 | 50.43 | 0.76 | 0.86 |
| | PA-EN | 32.19 | 56.84 | 0.64 | 60.3 | 0.799 | 0.82 |
| | EN-TA | 7.75 | 102.9 | 0.25 | 54.1 | 0.34 | 0.83 |
| | TA-EN | 24.44 | 66.86 | 0.57 | 54.91 | 0.74 | 0.84 |
| | EN-TE | 14.58 | 85.44 | 0.4 | 55.32 | 0.64 | 0.85 |
| | TE-EN | 29.32 | 63.04 | 0.62 | 59.83 | 0.76 | 0.83 |

Table 3 displays evaluation metrics for the models that use Sentencepiece for tokenization. The BLEU score for the NMT model, with Sentencepiece tokenization, ranges from 0.48 to 32.54. The METEOR score varies between 0.06 to 0.64, while the TER score falls between 63.48 and 146.82. The range of the Chrf and RIBES scores is 10.38 to 62.05 and 0.100 to 0.77, respectively. COMET score ranges are likewise between 0.54 and 0.85. However, when applying Sentencepiece tokenization for the Statistical MT model, the TER spans from 84.80 to 101.68 and the BLEU score falls between 1.37 and 12.39. The range of the METEOR score is 0.16 to 0.48, while the chrf score is 28.42 to 49.16. The COMET score ranges from 0.49 to 0.79, while the RIBES score is 0.08 to 0.61. In the same way, the ranges for BLEU and TER scores for MNMT models are 4.6 to 32.97 and 57.89 to 98.41, respectively. CHRF range is 35.76 to 62.35, and the METEOR score ranges from 0.23 to 0.65. The COMET score ranges are 0.71 and 0.87, while the RIBES score is 0.34 to 0.79.

Table 3: Evaluation Metrics using Sentencepiece as Tokenization

| Model | Language Pair | BLEU | TER | METEOR | CHRF | RIBES | COMET |
|---|---|---|---|---|---|---|---|
| SMT | EN-AS | 2.21 | 96.57 | 0.16 | 28.42 | 0.49 | 0.61 |
| | AS-EN | 3.77 | 87.8 | 0.27 | 32.34 | 0.47 | 0.7 |
| | EN-ML | 1.37 | 100.05 | 0.069 | 30.79 | 0.08 | 0.51 |
| | ML-EN | 2.02 | 91.45 | 0.15 | 25.59 | 0.34 | 0.68 |
| | EN-BN | 6.86 | 89.15 | 0.28 | 43.92 | 0.6 | 0.79 |
| | BN-EN | 13 | 84.8 | 0.48 | 49.16 | 0.59 | 0.8 |
| | EN-MR | 4.61 | 98.37 | 0.24 | 38.82 | 0.47 | 0.75 |
| | MR-EN | 9.75 | 85.35 | 0.42 | 45.57 | 0.55 | 0.63 |
| | EN-GU | 4.69 | 94.96 | 0.24 | 34.13 | 0.43 | 0.79 |
| | GU-EN | 11.63 | 83.63 | 0.46 | 48.48 | 0.58 | 0.77 |
| | EN-KN | 4.58 | 101.68 | 0.22 | 42.92 | 0.38 | 0.75 |
| | KN-EN | 10.05 | 88.33 | 0.42 | 46.15 | 0.51 | 0.76 |
| | EN-HI | 4.54 | 86.03 | 0.19 | 27.25 | 0.48 | 0.62 |
| | HI-EN | 4.93 | 87.95 | 0.22 | 34.61 | 0.46 | 0.57 |
| | EN-OR | 3.07 | 96.77 | 0.21 | 37.01 | 0.54 | 0.73 |
| | OR-EN | 7.21 | 86.45 | 0.35 | 41.18 | 0.51 | 0.74 |
| | EN-PA | 8 | 86.1 | 0.342 | 39.54 | 0.61 | 0.77 |
| | PA-EN | 12.39 | 80.24 | 0.46 | 47.65 | 0.58 | 0.75 |
| | EN-TE | 6 | 103.94 | 0.25 | 43.81 | 0.38 | 0.76 |
| | TE-EN | 11.39 | 87.4 | 0.46 | 49.06 | 0.53 | 0.76 |
| | EN-TA | 1.31 | 103.7 | 0.09 | 32.42 | 0.06 | 0.58 |
| | TA-EN | 2.6 | 90.12 | 0.18 | 28.11 | 0.38 | 0.72 |
| NMT | EN-AS | 0.48 | 146.82 | 0.06 | 10.38 | 0.1 | 0.54 |
| | AS-EN | 0.84 | 152.21 | 0.16 | 20.49 | 0.16 | 0.45 |
| | EN-ML | 8.7 | 99.91 | 0.28 | 53.43 | 0.45 | 0.84 |
| | ML-EN | 23.12 | 72.97 | 0.57 | 55.43 | 0.72 | 0.83 |
| | EN-BN | 17.43 | 74.33 | 0.43 | 53.47 | 0.72 | 0.86 |
| | BN-EN | 28.61 | 63.48 | 0.6 | 58.62 | 0.76 | 0.85 |
| | EN-MR | 9.88 | 95.03 | 0.33 | 44.23 | 0.59 | 0.82 |
| | MR-EN | 20.38 | 72.93 | 0.53 | 51.82 | 0.71 | 0.69 |

|  | EN-GU | 16.54 | 82.29 | 0.42 | 49.97 | 0.68 | 0.85 |
|---|---|---|---|---|---|---|---|
|  | GU-EN | 24.79 | 68.38 | 0.58 | 56.4 | 0.74 | 0.85 |
|  | EN-KN | 12.18 | 92.07 | 0.35 | 53.1 | 0.57 | 0.83 |
|  | KN-EN | 22.42 | 71.9 | 0.55 | 54.22 | 0.72 | 0.82 |
|  | EN-HI | 31.85 | 60.47 | 0.56 | 57.03 | 0.77 | 0.84 |
|  | HI-EN | 32.54 | 63.54 | 0.64 | 62.05 | 0.76 | 0.78 |
|  | EN-OR | 5.4 | 99.26 | 0.26 | 39.11 | 0.6 | 0.75 |
|  | OR-EN | 11.11 | 90.18 | 0.4 | 41.06 | 0.6 | 0.76 |
|  | EN-PA | 20.3 | 71.15 | 0.48 | 49.32 | 0.75 | 0.85 |
|  | PA-EN | 27.4 | 62.02 | 0.6 | 56.87 | 0.77 | 0.81 |
|  | EN-TE | 14.22 | 92.89 | 0.39 | 54.61 | 0.61 | 0.84 |
|  | TE-EN | 26.62 | 67.54 | 0.59 | 57.89 | 0.75 | 0.82 |
|  | EN-TA | 8.42 | 105.85 | 0.25 | 53.87 | 0.33 | 0.82 |
|  | TA-EN | 22.69 | 71.31 | 0.55 | 53.75 | 0.72 | 0.84 |
| MNMT | EN-AS | 4.6 | 98.41 | 0.23 | 35.76 | 0.59 | 0.8 |
|  | AS-EN | 19.9 | 75.19 | 0.51 | 50.4 | 0.7 | 0.78 |
|  | EN-ML | 8.44 | 96.78 | 0.29 | 53.95 | 0.48 | 0.85 |
|  | ML-EN | 26.68 | 66.55 | 0.59 | 57.67 | 0.74 | 0.84 |
|  | EN-BN | 15.22 | 76.59 | 0.41 | 52.18 | 0.71 | 0.86 |
|  | BN-EN | 27.93 | 63.66 | 0.6 | 58.02 | 0.76 | 0.85 |
|  | EN-MR | 9.92 | 92.17 | 0.35 | 46.84 | 0.63 | 0.84 |
|  | MR-EN | 25.89 | 65.34 | 0.59 | 56.83 | 0.75 | 0.71 |
|  | EN-GU | 17.68 | 77.42 | 0.45 | 50.96 | 0.71 | 0.87 |
|  | GU-EN | 29.85 | 61.04 | 0.63 | 59.99 | 0.77 | 0.86 |
|  | EN-KN | 11.77 | 87.91 | 0.35 | 52.21 | 0.59 | 0.84 |
|  | KN-EN | 24.83 | 68.37 | 0.57 | 55.8 | 0.73 | 0.83 |
|  | EN-HI | 28.4 | 61.55 | 0.54 | 55.2 | 0.77 | 0.86 |
|  | HI-EN | 32.97 | 57.79 | 0.65 | 62.35 | 0.79 | 0.79 |
|  | EN-OR | 9.99 | 83.12 | 0.35 | 47.13 | 0.68 | 0.85 |
|  | OR-EN | 25.74 | 67.05 | 0.59 | 56.86 | 0.74 | 0.8 |
|  | EN-PA | 19.98 | 69.94 | 0.49 | 49.43 | 0.76 | 0.86 |
|  | PA-EN | 31.2 | 57.89 | 0.63 | 59.85 | 0.79 | 0.82 |
|  | EN-TE | 13.25 | 87.9 | 0.39 | 54.15 | 0.62 | 0.84 |
|  | TE-EN | 28.66 | 64.83 | 0.61 | 59.43 | 0.75 | 0.83 |
|  | EN-TA | 7.6 | 102.04 | 0.25 | 54.41 | 0.34 | 0.83 |
|  | TA-EN | 24.59 | 67.87 | 0.56 | 54.79 | 0.73 | 0.85 |

Table 4 presents the outcomes of word piece tokenization using SMT, NMT, and MMMT. The performance metrics in SMT show a range of values. The range of the BLEU scores is 2.06 to 14.95. RIBES scores range from 0.08 to four. CHRF scores range from 27.13 to 51.82, while COMET scores fluctuate from 0.62 to 0.80. METEOR scores exhibit variability from 0.15 to 0.50, while TER scores range from 78.42 to 109.52. Similarly, the BLEU scores for NMT display a range of 0.48 to 31.95. The range of RIBES scores is 0.12 to 0.78. The range of CHRF scores is 12.22 to 61.45. COMET scores range from 0.52 to 0.86. The range of TER scores is 58.79 to 150.74. METEOR scores vary widely, ranging from 0.06 to 0.62. Likewise, the BLEU scores in MNMT showcase a wider range, ranging from 1.50 to 32.43. The range of RIBES scores is 0.28 to 0.78. The range of CHR F scores is 15.00 to 59.90. The range of COMET scores is 591 to 619. METEOR scores range from 129 to 391 whereas TER scores show variability from 119 to 319. After analyzing the NMT and SMT models, it was found that, for the majority of language pairs, the SentencePiece tokenizer consistently produced better results in terms of BLEU score than other tokenizers. However, it has been found that BPE tokenization performed better than other tokenization techniques in the context of MNMT.

Table 4: Evaluation Metrics using WordPiece as Tokenization

| Model | Language PPair | BLEU | TER | METEOR | CHRF | RIBES | COMET |
|---|---|---|---|---|---|---|---|
| SMT | EN-AS | 2.06 | 92.84 | 0.15 | 27.13 | 0.51 | 0.62 |
| | AS-EN | 3.83 | 88 | 0.26 | 32.07 | 0.47 | 0.69 |
| | EN-ML | 3.47 | 109.52 | 0.16 | 41.57 | 0.2 | 0.74 |
| | ML-EN | 7.95 | 85.27 | 0.37 | 43.81 | 0.46 | 0.77 |
| | EN-BN | 3.03 | 106.3 | 0.2 | 35.08 | 0.52 | 0.71 |
| | BN-EN | 6.64 | 84.77 | 0.32 | 38.86 | 0.51 | 0.73 |
| | EN-MR | 4.59 | 98.35 | 0.23 | 39.08 | 0.47 | 0.75 |
| | MR-EN | 9.75 | 85.67 | 0.41 | 45.63 | 0.54 | 0.63 |
| | EN-GU | 7.5 | 91.32 | 0.31 | 40.94 | 0.55 | 0.79 |
| | GU-EN | 11.45 | 84.82 | 0.46 | 48.35 | 0.58 | 0.81 |
| | EN-KN | 4.72 | 98.81 | 0.22 | 42.94 | 0.39 | 0.75 |
| | KN-EN | 9.72 | 89.08 | 0.42 | 46.02 | 0.51 | 0.76 |
| | EN-HI | 12.57 | 79.71 | 0.39 | 44.2 | 0.61 | 0.8 |
| | HI-EN | 14.95 | 78.42 | 0.5 | 51.82 | 0.61 | 0.7 |
| | EN-OR | 3.65 | 93.93 | 0.22 | 36.91 | 0.55 | 0.73 |
| | OR-EN | 7.39 | 85.33 | 0.37 | 41.18 | 0.52 | 0.74 |
| | EN-PA | 6.84 | 82.1 | 0.31 | 39.54 | 0.61 | 0.77 |
| | PA-EN | 10.39 | 80.14 | 0.32 | 47.65 | 0.58 | 0.75 |
| | EN-TE | 5.74 | 101.42 | 0.25 | 44.19 | 0.4 | 0.762 |
| | TE-EN | 5.74 | 101.42 | 0.25 | 44.19 | 0.4 | 0.76 |

|  | Pair | | | | | | |
|---|---|---|---|---|---|---|---|
|  | EN-TA | 3.59 | 114.8 | 0.16 | 46.68 | 0.13 | 0.73 |
|  | TA-EN | 9.21 | 89.17 | 0.39 | 43.74 | 0.5 | 0.79 |
| NMT | EN-AS | 0.48 | 133.12 | 0.063 | 12.22 | 0.12 | 0.52 |
|  | AS-EN | 0.75 | 150.74 | 0.15 | 19.69 | 0.15 | 0.54 |
|  | EN-ML | 8.76 | 101.77 | 0.28 | 53.8 | 0.44 | 0.83 |
|  | ML-EN | 22.74 | 72.06 | 0.55 | 54.71 | 0.71 | 0.83 |
|  | EN-BN | 6.62 | 86.61 | 0.24 | 43.08 | 0.63 | 0.54 |
|  | BN-EN | 0.42 | 128.66 | 0.14 | 20.23 | 0.15 | 0.8 |
|  | EN-MR | 9.37 | 94.91 | 0.32 | 44.73 | 0.58 | 0.81 |
|  | MR-EN | 20.32 | 72.96 | 0.52 | 51.9 | 0.71 | 0.69 |
|  | EN-GU | 16.56 | 80.68 | 0.42 | 50.05 | 0.685 | 0.85 |
|  | GU-EN | 24.33 | 68.23 | 0.57 | 56.27 | 0.74 | 0.85 |
|  | EN-KN | 12.96 | 85.69 | 0.34 | 53.44 | 0.59 | 0.83 |
|  | KN-EN | 21.63 | 73.46 | 0.54 | 54.11 | 0.71 | 0.83 |
|  | EN-HI | 31.19 | 58.79 | 0.55 | 56.75 | 0.78 | 0.86 |
|  | HI-EN | 31.95 | 59.57 | 0.62 | 61.45 | 0.78 | 0.79 |
|  | EN-OR | 4.34 | 105.44 | 0.24 | 36.98 | 0.59 | 0.73 |
|  | OR-EN | 9.1 | 93 | 0.36 | 38.235 | 0.55 | 0.74 |
|  | EN-PA | 19.43 | 71.04 | 0.47 | 49.34 | 0.74 | 0.84 |
|  | PA-EN | 27.15 | 63.18 | 0.59 | 56.96 | 0.77 | 0.81 |
|  | EN-TE | 13.77 | 89.12 | 0.38 | 54.28 | 0.61 | 0.83 |
|  | TE-EN | 25.28 | 68.98 | 0.57 | 56.91 | 0.72 | 0.82 |
| MNMT | EN-AS | 1.5 | 103.8 | 0.08 | 15 | 0.28 | 0.78 |
|  | AS-EN | 19 | 74.39 | 0.47 | 47.98 | 0.69 | 0.57 |
|  | EN-ML | 9.16 | 94.64 | 0.27 | 51.97 | 0.46 | 0.85 |
|  | ML-EN | 26.02 | 66.18 | 0.58 | 57.4 | 0.74 | 0.83 |
|  | EN-BN | 15.35 | 74.63 | 0.4 | 50.79 | 0.72 | 0.85 |
|  | BN-EN | 26.86 | 65.18 | 0.58 | 57.73 | 0.75 | 0.84 |
|  | EN-MR | 10.96 | 87.03 | 0.35 | 46.29 | 0.64 | 0.81 |
|  | MR-EN | 26.23 | 64.23 | 0.58 | 57.02 | 0.75 | 0.71 |
|  | EN-GU | 17.55 | 76.16 | 0.43 | 49.55 | 0.7 | 0.87 |
|  | GU-EN | 30.28 | 59.82 | 0.62 | 60.53 | 0.78 | 0.86 |
|  | EN-KN | 11.28 | 86.28 | 0.33 | 50.99 | 0.58 | 0.84 |
|  | KN-EN | 24.15 | 68.11 | 0.56 | 55.77 | 0.74 | 0.82 |
|  | EN-HI | 27.25 | 61.7 | 0.52 | 53.23 | 0.77 | 0.86 |
|  | HI-EN | 32.43 | 58.04 | 0.63 | 62.07 | 0.79 | 0.78 |
|  | EN-OR | 9.5 | 81.2 | 0.32 | 45.74 | 0.68 | 0.81 |
|  | OR-EN | 25.05 | 66.45 | 0.573 | 55.84 | 0.73 | 0.8 |

|  | EN-PA | 19.48 | 68.7 | 0.47 | 48.3 | 0.75 | 0.86 |
|---|---|---|---|---|---|---|---|
|  | PA-EN | 30.78 | 57.92 | 0.62 | 59.79 | 0.79 | 0.81 |
|  | EN-TE | 14.22 | 84.03 | 0.37 | 52.61 | 0.62 | 0.84 |
|  | TE-EN | 28.8 | 63.73 | 0.6 | 59.9 | 0.76 | 0.82 |
|  | EN-TA | 7.56 | 101.61 | 0.24 | 52.09 | 0.3 | 0.83 |
|  | TA-EN | 23.12 | 68.15 | 0.55 | 53.97 | 0.73 | 0.841 |

## 6. Conclusion

This paper aims at investigating the effects of various subword tokenization techniques, such as Byte Pair Encoding (BPE), SentencePiece, and WordPiece, on Indian languages. It has been found that SentencePiece tokenization outperforms other tokenizers in both Statistical Machine Translation (SMT) and Neural Machine Translation (NMT) tasks, according to the results of 11 IL experiments. However, BPE tokenization outperforms Multilingual Neural Machine Translation (MNMT). This difference in efficiency can be attributed to the nature of the tasks as well as the linguistic characteristics of the languages used. SentencePiece is particularly well-suited for SMT and NMT tasks due to its ability to capture morphological nuances and handle out-of-vocabulary words. However, in the context of MNMT, BPE's segmentation strategy effectively represents subword units proving advantageous. The results show that, despite using different tokenizers on the Samanantar dataset for each model (SMT, NMT, and MNMT), translations from ILs to English outperform translations from English to ILs. In future work, this tokenizer can be used to translate Indic to Indic languages and the translation quality can be evaluated further.